\documentclass[twoside,twocolumn,12pt]{article}

\usepackage{array}
\pdfoutput=1

\usepackage{microtype} 

\usepackage[hmarginratio=1:1,top=25mm,columnsep=20pt]{geometry} 
\usepackage[hang, small,labelfont=bf,up,textfont=it,up]{caption} 
\usepackage{booktabs} 
\usepackage{natbib}

\usepackage{enumitem} 
\setlist[itemize]{noitemsep} 

\usepackage{abstract} 

\usepackage{titlesec} 
\renewcommand\thesection{\Roman{section}} 
\renewcommand\thesubsection{\roman{subsection}} 
\titleformat{\section}[block]{\large\scshape\centering}{\thesection.}{1em}{} 
\titleformat{\subsection}[block]{\large}{\thesubsection.}{1em}{} 

\usepackage{titling} 

\usepackage{hyperref} 

\usepackage{graphicx}
\graphicspath{ {./images/} }

\usepackage{xurl}

\pretitle{\begin{center}\Huge} 
\posttitle{\end{center}} 
\title{Extractive Question Answering on Queries in Hindi and Tamil} 
\author{Adhitya Thirumala, Elisa Ferracane
}
\date{}

\renewcommand{\maketitlehookd}{%
\vspace{-3.75em}
\begin{abstract}
\vspace{-1em}
\noindent Indic languages like Hindi and Tamil are underrepresented in the natural language processing (NLP) field compared to languages like English. Due to this underrepresentation, performance on NLP tasks (such as search algorithms) in Indic languages are inferior to their English counterparts. This difference disproportionately affects those who come from lower socioeconomic statuses because they consume the most Internet content in local languages. The goal of this project is to build an NLP model that performs better than pre-existing models for the task of extractive question-answering (QA) on a public dataset in Hindi and Tamil. Extractive QA is an NLP task where answers to questions are extracted from a corresponding body of text. To build the best solution, we used three different models. The first model is an unmodified cross-lingual version of the NLP model RoBERTa, known as XLM-RoBERTa, that is pretrained on 100 languages. The second model is based on the pretrained RoBERTa model with an extra classification head for the question answering, but we used a custom Indic tokenizer, then optimized hyperparameters and fine tuned on the Indic dataset. The third model is based on XLM-RoBERTa, but with extra finetuning and training on the Indic dataset. We hypothesize the third model will perform best because of the variety of languages the XLM-RoBERTa model has been pretrained on and the additional finetuning on the Indic dataset. This hypothesis was proven wrong because the paired RoBERTa models performed the best as the training data used was most specific to the task performed as opposed to the XLM-RoBERTa models which had much data that was not in either Hindi or Tamil.
\end{abstract}
}

\begin{document}

\maketitle


\section{Introduction}

 Indic languages are underrepresented in the natural language processing (NLP) field compared to languages like English. For example, in the corpora provided by the Natural Language Toolkit (NLTK) (\cite{arora-2020-inltk}), over half of them are in English. Indic languages include Hindi, Tamil, Telugu, Malayalam, and others that originate from the Indian subcontinent, and are spoken by the 1.7 billion people that live on the Indian subcontinent. As seen in Table 1, around 57\% speak Hindi, while only 10.6\% speak English. An English-centric view is bad for the NLP community for a variety of reasons. First, English’s written script is different from other languages, and a focus on English or only English-like languages (such as French, German, Spanish, etc.) can make current systems work poorly on languages with different writing systems, such as Chinese or Korean. Second, tokenization, which is the process of splitting a document into smaller pieces, such as paragraphs, sentences, or words, is more straightforward in English and other languages that have an easier way to define a “word.” However, in languages such as Tamil where one does not traditionally put spaces between words, the task of tokenizing changes and becomes completely different (\cite{bender_2021}).  Due to this under-representation, NLP tasks in Indic languages, such as search algorithms and sentiment analysis,  have performance inferior to their English counterparts (\cite{arora-2020-inltk}). This difference disproportionately affects those who come from lower socioeconomic statuses because they consume the most Internet content in local languages (\cite{s_2019}). One way to allow people to access information easier is to improve the performance of extractive question-answering algorithms on search engines in Indic languages. Extractive QA is an NLP task that extracts an answer to questions from a given body of text. This body of text is known as the context and normally contains the answer to the question provided. In the dataset we used, the contexts (and answers) came from Wikipedia articles, and the questions were generated by native speakers of both Hindi and Tamil. For this project, we will use Hindi and Tamil as train data as they are the ones provided by the Kaggle competition (\cite{kaggle_comp}) that inspired this project. 

\begin{table}
\caption{Total speakers of languages in India \scriptsize (Source: 2011 Indian Census)}
\centering
\begin{tabular}{m{2.5cm} m{2cm} m{1.5cm}  } 
\toprule

Language & Speakers(M) & Speaker\% in Pop. \\
\midrule
Hindi & 692 & 57.1 \\

English & 129 & 10.6 \\

Bengali & 107 & 8.9 \\

Marathi & 99 & 8.2 \\

Telugu & 95 & 7.8 \\

Tamil & 77 & 6.3 \\

Gujarati & 60 & 5 \\

Urdu & 63 & 5.2 \\

Kannada & 59 & 4.9 \\

Malayalam & 36 & 2.9 \\

Punjabi & 36 & 3 \\

Assamese & 24 & 2 \\

Maithili & 14 & 1.2 \\

Sanskrit & 0.025 & 0.3 \\
\bottomrule

\end{tabular}
\end{table}

\section{Related Work}
Previously, the task of extractive QA on English datasets was popularized in SQuAD: 100,000+ Questions for Machine Comprehension of Text (\cite{squad_v1}). This paper outlined the creation process for an English dataset for the task of Extractive QA. The questions were proposed by human annotators and based on Wikipedia articles. Similarly, the dataset that we used was also derived from Wikipedia articles and had questions produced by human annotators. However, our dataset is different because it is one of the first datasets produced in Hindi and Tamil. Currently, in English, the best models on extractive question answering use transformer models. Transformers are a computationally efficient deep learning model that introduce attention which allows the model to give more or less importance to different parts of the input. This mechanism is particularly useful in the task of question answering because it allows the model to ignore the useless parts of the context and obtain the answer from the useful parts. We utilize transformer models such as XLM-RoBERTa and RoBERTa to achieve the highest performance on the task. We describe the changes we made to these pre-trained models in Section VI.
\section{Task}
Extractive question answering is the task of extracting the question to the answer from a corresponding context. The inputs for this task are a question, and a body of text known as the context. This body of text generally contains the answer; however, it does not have to. Certain datasets such as SQuADv2 (\cite{squad_v2}) include questions with no answer provided, but the dataset that we use does not do this. The output for this task is a span of text in the context or a null value if the answer does not exist in the context. This output is presented to the end-user as the answer to the question.
\section{Evaluation Metric}
We utilize the word-level Jaccard score for this task which tests text similarity between two strings. The python implementation for this method is below.
\begin{small}
\begin{verbatim}
def jaccard(str1, str2): 
    a = set(str1.lower().split()) 
    b = set(str2.lower().split())
    c = a.intersection(b)
    d = (len(a) + len(b) - len(c))
    return float(len(c)) / d
\end{verbatim}
\end{small}
This function takes the predicted answer and correct answer as inputs. Then, it defines 3 variables, known as a, b, and c. a and b are the correct answer and predicted answer respectively split into a list of characters. c is the common characters between both a and b, and d is the total number of characters in the inputs subtracted by the number of common characters. The function returns the number of common characters between both inputs divided by d. The equation to apply this metric is defined as such.

\[ \frac{1}{n} \sum_{i=1}^{n} jaccard(answer_{predicted}, answer_{correct})\]
In this equation, $n$ is the number of predictions made, $jaccard$ is the function defined above, $answer_{predicted}$ is the predicted answer and $answer_{correct}$ is the correct answer. A Jaccard score of 0 means that none of the characters are the same between the predicted answer and the correct answer, and a score of 1 means that the two are identical. The implementation of this metric and the equation above are from the kaggle competition (\cite{kaggle_comp}).
\section{Data}

For this project, all of our data came from the Kaggle competition called chaii - Hindi and Tamil Question Answering (\cite{kaggle_comp}). In this dataset, the context, question, answer text, the position of the starting word of the answer, and language of the question are provided. Around 23\% of the dataset (740 entries) is in Hindi while around 13\% of the dataset (364 entries) is in Tamil. The entire dataset contains 1104 entries. The creators of the Kaggle competition created the dataset in two steps with help from native speakers of both Hindi and Tamil. First, in the question elicitation step, the native speakers were shown excerpts from Wikipedia articles in their respective languages. Based on these articles, participants were asked to create questions about the excerpts that they were genuinely interested in knowing the answer to. These questions did not have an answer that was in the Wikipedia article verbatim. After this, in the second step of answer labeling, more annotators were shown the questions from the previous step and the first Wikipedia article that the Google search for the question produced. The Wikipedia article selected was to be in Hindi for Hindi questions and Tamil for Tamil questions. The annotators were then asked to provide the first answer to the question shown in the Wikipedia article. If the answer to the question could not be found in this Wikipedia article, the question was thrown out and not included in the final dataset. For the training dataset, 1 annotator did the steps shown here. For the testing dataset, 3 annotators did the steps and a voting process determined the final answer. If there were disagreements, a separate pass was done to determine the final answer. This method may lead to some noise, however, this is representative of real-world scenarios and will lead to a more noise-tolerant model. This method was created in TydiQA (\cite{tydiqa}). Our validation dataset was a random 64 entries selected from the training set.  
\section{Models}
We evaluated the Hindi and Tamil extractive QA task on 4 models. The first two are based on XLM-RoBERTa and the last two are based on RoBERTa. To import all of these models into our code, we used a package called transformers which was provided by the NLP company Huggingface (\cite{huggingface}). A transformer is a computationally efficient deep learning model that introduces attention which allows the model to give more or less importance to different parts of the input. It does not use convolutions. These transformers are pre-trained models that can be used by Tensorflow or PyTorch depending on the model selected. For this project, we obtained the pre-trained XLM-RoBERTa model from the python package Transformers and the Indic version of RoBERTa from a Kaggle post. We next describe each model in detail.
\subsection{XLM-RoBERTa} We first used an off-the-shelf version of XLM-RoBERTa (\cite{xlm}). XLM-RoBERTa is a fork of RoBERTa which can work on any language. This is achieved by training the model on a large corpus of webcrawl data called the Common Crawl. The Common Crawl is a dataset that contains over 7 years of webcrawl data in over 40 languages. The question answering head applied to XLM-RoBERTa was trained on the Stanford Question Answering Dataset (SQuAD) (\cite{squad_v2}).  This model was used as a baseline and was not trained on any Tamil or Hindi data. 
\subsection{XLM-RoBERTa+finetune} Our second model is based on XLM-RoBERTa, but further trained on both the Hindi and Tamil data to gain further performance. We train for 2 epochs with a learning rate of 2 x 10\textsuperscript{-5}. 
\subsection{RoBERTa + Hindi finetune/Tamil finetune} Our third and fourth models are based on RoBERTa (\cite{roberta}). RoBERTa is a better trained version of BERT (\cite{bert}) which trained with better hyperparameters and achieves higher NLP performance on all tasks. Since RoBERTa is a model trained purely on English, we had to obtain a tokenizer that was capable of tokenizing Hindi and Tamil text. To achieve this, we obtained a version of RoBERTa that was modified by WECHSEL (\cite{wechsel}). WECHSEL is a python package that initializes subword embeddings to transfer monolingual NLP models into different languages. For the purpose of this project, we obtained 2 versions of RoBERTa that were transferred into Hindi and Tamil respectively. Both of these models were trained exclusively on Hindi (RoBERTa+Hindi finetune) and Tamil (RoBERTa+Tamil finetune) QA data provided by the Kaggle competition. The hyperparameters for these models are identical. We trained for 4 epochs with batch size 12, learning rate 2 x 10\textsuperscript{-5}, and weight decay 0.01. 
\section{Results}
In this section, we describe the results of our models.

\subsection{XLM-RoBERTa}
For the XLM-RoBERTa model, we achieved a word-level Jaccard score of 0.656 on both the Hindi and Tamil dataset combined. This was on the validation set. For this project, this will serve as the baseline model that all other models are compared to.
\subsection{XLM-RoBERTa+finetune}
For the XLM-RoBERTa+finetune model, we achieved a word-level Jaccard score of 0.749 on both the Hindi and Tamil datasets combined. This prediction was also run on the same 64 examples from the previous section. This is an increase of 14.18\% from the XLM-RoBERTa model. 
\subsection{RoBERTa + Hindi finetune/Tamil finetune}
The best performing model was the RoBERTa+ Hindi finetune/Tamil finetune model. we achieved a Jaccard score of 0.958 with the Hindi model. We achieved a Jaccard score of 0.829 with the Tamil model. The reason that these scores are different is that there the dataset that is used is 67\% Hindi and 33\% Tamil. The average of these two scores is 0.893. This is an increase of 36.13\% over the XLM-RoBERTa model and a increase of 19.23\% over the XLM-RoBERTa+finetune model.
 \begin{figure}   
   \caption{Incorrect Model Outputs and Corresponding Answers}
   \vspace{-1.2em}
   \begin{center}
     \includegraphics[scale=0.4]{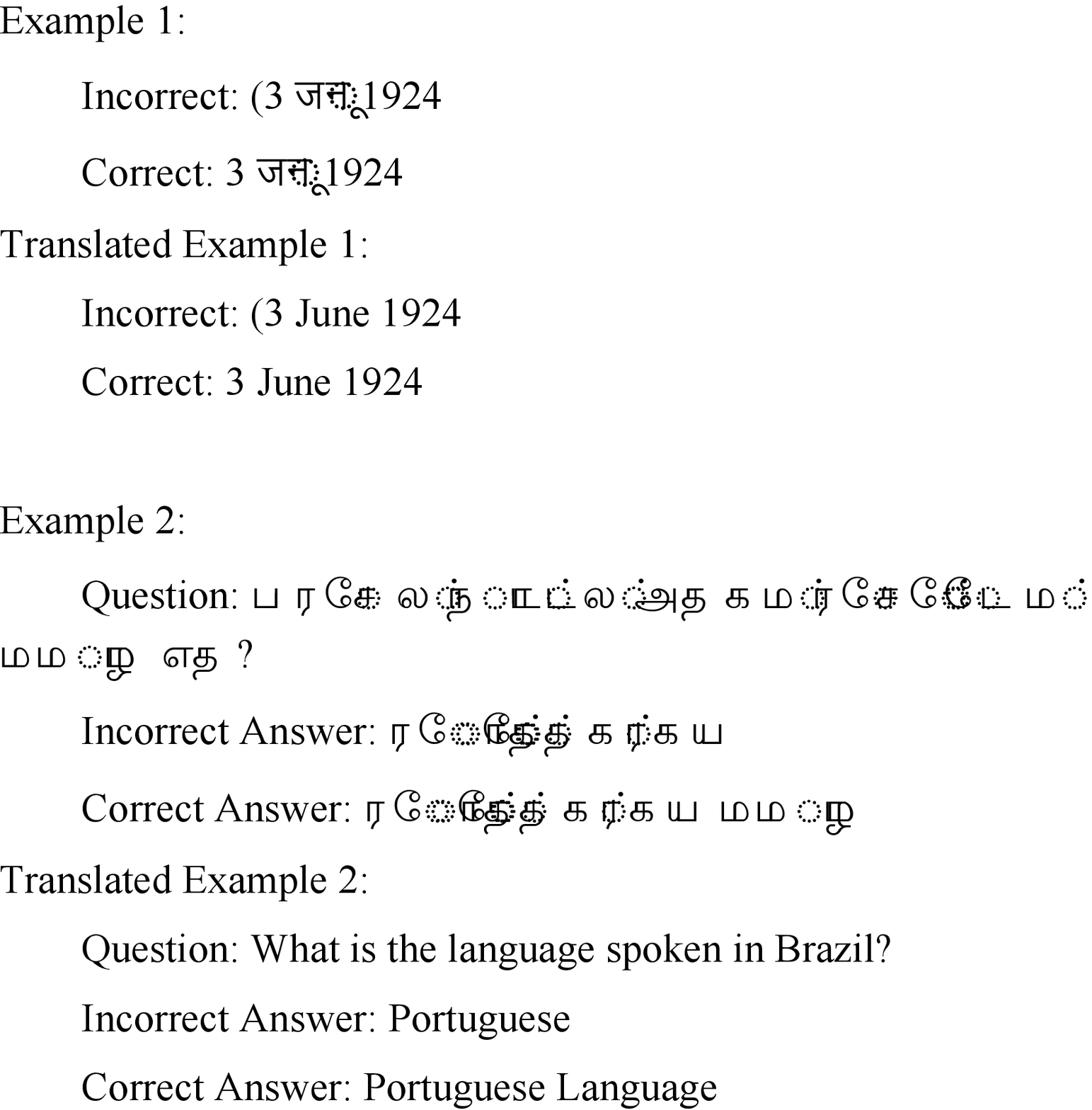}
   \end{center}
   \vspace{-1.5em}
\end{figure}
\section{Error Analysis}
The RoBERTa model performed much better than both of the XLM-RoBERTa models. This could be due to the fact that the data fed to the RoBERTa models is more specific to the task that the model is performing. The XLM models must use both Tamil and Hindi data for both languages while the RoBERTa model uses only Tamil data for Tamil predictions and only Hindi data for Hindi predictions. The models made some errors in predictions and most of them can be explained reasonably. The first example can be seen in Example 1. Incorrectly, the XLM model picked up the parenthesis as part of the answer. While the predicted answer is functionally equivalent to the correct answer, the Jaccard score will still be lower. It is preferable to remove these kinds of errors for user accessibility and UI cleanliness. The next error can be seen in Example 2. The question given here can be translated as, ``What is the language spoken in Brazil?'' The model output is translated as ``Portuguese.'' However, in Tamil, the correct answer would be translated as ``Portuguese Language.'' This error was also in the XLM model and could be because the model was trained on English grammar. In English grammar, the answer given would be correct. This is a major reason to continue the development of NLP algorithms in languages other than English to avoid situations like these.

\section{Conclusion}
In this project, we approached the task of extractive question answering on queries in Hindi and Tamil. This problem is important because most NLP systems are based in English and an English-centric view in NLP can cause models in other languages to perform worse as English functions differently from many languages. We used 3 models to achieve our goal. The first model was the XLM-RoBERTa. This model was a baseline model and achieved a word-level Jaccard score of 0.656. The next model was the XLM-RoBERTa+finetune. This model achieved a word-level Jaccard score of 0.749. The last and best performing model was RoBERTa+ Hindi finetune/Tamil finetune. This model achieved a word-level Jaccard score of 0.958 in Hindi and a score of 0.829 in Tamil. The biggest application of this project is to improve extractive QA models in search engines for different languages. This will help those who don't know English access information far easier. Any code for this project can be found by clicking \href{https://github.com/adhi-thirumala/Science-Fair2022}{here}.

\section{Acknowledgments}
 We would like to thank Dr. Anita Balachandran, for helping analyze the outputs of our model since she can read both Hindi and Tamil.

\bibliographystyle{plainnat}
\bibliography{bibliography}

\begin{thebibliography}{12}
\providecommand{\natexlab}[1]{#1}
\providecommand{\url}[1]{\texttt{#1}}
\expandafter\ifx\csname urlstyle\endcsname\relax
  \providecommand{\doi}[1]{doi: #1}\else
  \providecommand{\doi}{doi: \begingroup \urlstyle{rm}\Url}\fi

\bibitem[Arora(2020)]{arora-2020-inltk}
Gaurav Arora.
\newblock i{NLTK}: Natural language toolkit for indic languages.
\newblock In \emph{Proceedings of Second Workshop for NLP Open Source Software
  (NLP-OSS)}, pages 66--71, Online, November 2020. Association for
  Computational Linguistics.
\newblock \doi{10.18653/v1/2020.nlposs-1.10}.
\newblock URL \url{https://www.aclweb.org/anthology/2020.nlposs-1.10}.

\bibitem[Bender(2021)]{bender_2021}
Emily~M. Bender.
\newblock The benderrule: On naming the languages we study and why it matters,
  Dec 2021.
\newblock URL
  \url{https://thegradient.pub/the-benderrule-on-naming-the-languages-we-study-and-why-it-matters/}.

\bibitem[Clark et~al.(2020)Clark, Choi, Collins, Garrette, Kwiatkowski,
  Nikolaev, and Palomaki]{tydiqa}
Jonathan~H. Clark, Eunsol Choi, Michael Collins, Dan Garrette, Tom Kwiatkowski,
  Vitaly Nikolaev, and Jennimaria Palomaki.
\newblock Tydi qa: A benchmark for information-seeking question answering in
  typologically diverse languages.
\newblock \emph{Transactions of the Association for Computational Linguistics},
  2020.

\bibitem[Conneau et~al.(2019)Conneau, Khandelwal, Goyal, Chaudhary, Wenzek,
  Guzm{\'{a}}n, Grave, Ott, Zettlemoyer, and Stoyanov]{xlm}
Alexis Conneau, Kartikay Khandelwal, Naman Goyal, Vishrav Chaudhary, Guillaume
  Wenzek, Francisco Guzm{\'{a}}n, Edouard Grave, Myle Ott, Luke Zettlemoyer,
  and Veselin Stoyanov.
\newblock Unsupervised cross-lingual representation learning at scale.
\newblock \emph{CoRR}, abs/1911.02116, 2019.
\newblock URL \url{http://arxiv.org/abs/1911.02116}.

\bibitem[Devlin et~al.(2018)Devlin, Chang, Lee, and Toutanova]{bert}
Jacob Devlin, Ming{-}Wei Chang, Kenton Lee, and Kristina Toutanova.
\newblock {BERT:} pre-training of deep bidirectional transformers for language
  understanding.
\newblock \emph{CoRR}, abs/1810.04805, 2018.
\newblock URL \url{http://arxiv.org/abs/1810.04805}.

\bibitem[Google(2021)]{kaggle_comp}
Google.
\newblock chaii - hindi and tamil question answering, July 2021.
\newblock URL
  \url{https://www.kaggle.com/c/chaii-hindi-and-tamil-question-answering}.

\bibitem[Liu et~al.(2019)Liu, Ott, Goyal, Du, Joshi, Chen, Levy, Lewis,
  Zettlemoyer, and Stoyanov]{roberta}
Yinhan Liu, Myle Ott, Naman Goyal, Jingfei Du, Mandar Joshi, Danqi Chen, Omer
  Levy, Mike Lewis, Luke Zettlemoyer, and Veselin Stoyanov.
\newblock Roberta: {A} robustly optimized {BERT} pretraining approach.
\newblock \emph{CoRR}, abs/1907.11692, 2019.
\newblock URL \url{http://arxiv.org/abs/1907.11692}.

\bibitem[Minixhofer et~al.(2021)Minixhofer, Paischer, and Rekabsaz]{wechsel}
Benjamin Minixhofer, Fabian Paischer, and Navid Rekabsaz.
\newblock {WECHSEL:} effective initialization of subword embeddings for
  cross-lingual transfer of monolingual language models.
\newblock \emph{CoRR}, abs/2112.06598, 2021.
\newblock URL \url{https://arxiv.org/abs/2112.06598}.

\bibitem[Rajpurkar et~al.(2016)Rajpurkar, Zhang, Lopyrev, and Liang]{squad_v1}
Pranav Rajpurkar, Jian Zhang, Konstantin Lopyrev, and Percy Liang.
\newblock Squad: 100, 000+ questions for machine comprehension of text.
\newblock \emph{CoRR}, abs/1606.05250, 2016.
\newblock URL \url{http://arxiv.org/abs/1606.05250}.

\bibitem[Rajpurkar et~al.(2018)Rajpurkar, Jia, and Liang]{squad_v2}
Pranav Rajpurkar, Robin Jia, and Percy Liang.
\newblock Know what you don't know: Unanswerable questions for squad.
\newblock \emph{CoRR}, abs/1806.03822, 2018.
\newblock URL \url{http://arxiv.org/abs/1806.03822}.

\bibitem[S(2019)]{s_2019}
Rukmini S.
\newblock In india, who speaks in english, and where?, May 2019.
\newblock URL
  \url{https://www.livemint.com/news/india/in-india-who-speaks-in-english-and-where-1557814101428.html}.

\bibitem[Wolf et~al.(2019)Wolf, Debut, Sanh, Chaumond, Delangue, Moi, Cistac,
  Rault, Louf, Funtowicz, and Brew]{huggingface}
Thomas Wolf, Lysandre Debut, Victor Sanh, Julien Chaumond, Clement Delangue,
  Anthony Moi, Pierric Cistac, Tim Rault, R{\'{e}}mi Louf, Morgan Funtowicz,
  and Jamie Brew.
\newblock Huggingface's transformers: State-of-the-art natural language
  processing.
\newblock \emph{CoRR}, abs/1910.03771, 2019.
\newblock URL \url{http://arxiv.org/abs/1910.03771}.

\end{thebibliography}


\end{document}